\pdfoutput=1

\documentclass[11pt]{article}

\usepackage{emnlp2021} %

\usepackage{times}
\usepackage{latexsym}

\usepackage[T1]{fontenc}

\usepackage[utf8]{inputenc}

\usepackage{microtype}

\usepackage{times}
\usepackage{latexsym}

\usepackage{enumerate}
\usepackage{amsmath}
\usepackage{amssymb}
\usepackage{graphicx}
\usepackage{pifont}
\usepackage{subcaption}
\usepackage{booktabs}
\usepackage{multirow}
\newcommand{\cmark}{\ding{51}}%
\newcommand{\xmark}{\ding{55}}%
\usepackage{dirtytalk}

\usepackage{microtype}

\title{Does It Capture STEL? A Modular, Similarity-based Linguistic Style Evaluation Framework}

\author{Anna Wegmann and Dong Nguyen \\
  Department of Information and Computing Sciences \\
  Utrecht University \\ %
  Utrecht, the Netherlands \\
  \texttt{a.m.wegmann@uu.nl}, \texttt{d.p.nguyen@uu.nl} %
  }

\begin{document}
\maketitle
\begin{abstract}
Style is an integral part of natural language. However, evaluation methods for style measures are rare, often task-specific and usually do not control for content. We propose the modular, fine-grained and content-controlled  \textit{similarity-based STyle  EvaLuation framework} (STEL) to test the performance of any model that can compare two sentences on style. We illustrate STEL with two general \textit{dimensions} of style (formal/informal and simple/complex) as well as two specific \textit{characteristics} of style (contrac'tion and numb3r substitution). We find that BERT-based methods outperform simple versions of commonly used style measures like 3-grams, punctuation frequency and LIWC-based approaches. We invite the addition of further tasks and task instances to STEL and hope to facilitate the improvement of style-sensitive measures.
\end{abstract}

\section{Introduction}

Natural language is not only about what is said (i.e., content), but also about how it is said (i.e., \textit{linguistic style}). Linguistic style and social context are highly interrelated \cite{coupland2007style, Bell2013TheGT}. %
For example, people can accommodate their linguistic style to each other based on %
social power differences \cite{danescu_echoes-power}. 
Furthermore, linguistic style can influence perception, e.g., the persuasiveness of news \cite{el-baff-etal-2020-analyzing} or the success of pitches on crowdsourcing platforms \cite{PARHANKANGAS2017215}.
As a result, style is relevant for natural language understanding, e.g., in author profiling \cite{10.1145/1871985.1871993}, abuse detection
\cite{markov-etal-2021-exploring} %
or understanding conversational interactions \cite{danescu-niculescu-mizil-lee-2011-chameleons}. 
Additionally, style can be important to address in natural language generation \cite{ficler-goldberg-2017-controlling}, including identity modeling in dialogue systems \cite{li-etal-2016-persona} and style preservation in machine translation \cite{niu-etal-2017-study, rabinovich-etal-2017-personalized}.

\begin{figure}[t]
    \centering
    \includegraphics[width=0.49\textwidth]{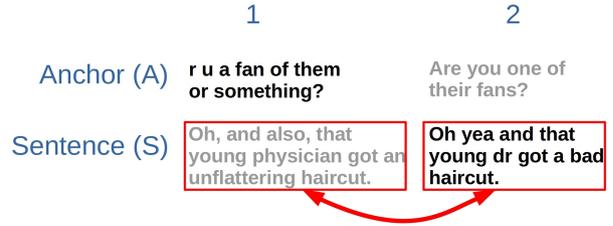}
    \caption{\textbf{STEL Task Instance.} Anchor 1 (A1) and anchor 2 (A2) and the alternative sentences 1 (S1) and 2 (S2) are split along the same style dimension (here: formal/informal). The sentences and anchors are paraphrases of each other. 
    The STEL task is to order S1 and S2 to match A1-A2.
    Here, the correct order is S2-S1.
    }
    \label{fig:STEL-Task}
\end{figure}

There are several general evaluation benchmarks for different linguistic phenomena (e.g., \newcite{wang-etal-2018-glue, NEURIPS2019_4496bf24}) but less emphasis has been put on linguistic style. 
Nevertheless, natural language processing literature shows a variety of  approaches for the evaluation of style measuring methods: 
They have been tested on whether they group texts by the same authors together \cite{hay-etal-2020-representation, PAN_2020}, whether they can correctly classify the style for
ground truth datasets \cite{niu-carpuat-2017-discovering, kang-hovy-2021-style}
and whether `similar style words' are similarly represented \cite{akama-etal-2018-unsupervised}. 
However, these evaluation approaches are (i) often application-specific, (ii) rarely used to compare different style methods, (iii) usually do not control for content and (iv) often do not test for fine-grained style differences.

These shortcomings (i)-(iv) might be the result of 
the following challenges for the construction of style evaluation methods: 1. Style is a highly ambiguous and elusive term \cite{biber_conrad_2009, crystal_english_style, labov_2006, xu-2017-shakespeare}. We propose a \textit{modular} framework where components can 
be removed or added to fit an application or specific understanding of style.  2. Variation in style can be very small. 
Our proposed evaluation framework can be used to test for \textit{fine-grained} style differences. %
3. Style is hard to disentangle from content as the two are often correlated  (e.g., \newcite{gero-etal-2019-low,bischoff2020importance}). For example, people might speak more formally in a job interview than in a bar with friends. Thus, language models and methods might pick up on spurious content correlations (similar to \newcite{poliak-etal-2018-hypothesis}) in a benchmark that does not \textit{control} for \textit{content}.

To this end, we propose the \textbf{modular, fine-grained and content-controlled  \textit{similarity-based STyle EvaLuation framework} (STEL)}. We demonstrate it for the English language. An example task is shown in Figure \ref{fig:STEL-Task}.
The task is to order sentence 1 (S1) and sentence 2 (S2) to match the style order of anchor 1 (A1) and anchor 2 (A2).  Our STEL framework encompasses two general \textit{dimensions} of style (formal/informal and simple/complex) as well as two specific  \textit{characteristics} of style (contraction and number substitution). By design, the style characteristics are easy to identify. Thus, the STEL characteristic tasks are easier to solve than the STEL dimension tasks. STEL contains 815 task instances per dimension and 100 task instances per characteristic (see Table \ref{table:quad-examples}). 
To be evaluated on STEL, style measuring methods need not be able to classify styles directly. Instead, any method that can calculate the style \textit{similarity} between two sentences can be evaluated: (1)  Style (measuring) methods that calculate similarity values directly (e.g., edit distance or cross-encoders \newcite{reimers-gurevych-2019-sentence}) and (2) vector representations of a sentence's style (e.g., \newcite{hay-etal-2020-representation, stylometric_rep_AA}) by using a distance or similarity measure between them (e.g., cosine similarity). This similarity-based setup also simplifies task extension (c.f.~modularity). %
STEL components can easily be generated from parallel sets of paraphrases which differ along a style dimension (\S \ref{sec:STEL-task}), e.g., sets of paraphrases that vary along the formal/informal dimension \cite{rao-tetreault-2018-dear}.

\paragraph*{Contribution.}
With this paper, we contribute \textbf{(a)} the modular, fine-grained and content-controlled  STEL framework (\S \ref{sec:stel-framework}), \textbf{(b)} 1830 validated task instances for the considered style components (\S \ref{sec:task-filtering}) and \textbf{(c)} baseline results of STEL on 18 style measuring methods (\S \ref{sec:ex-setup}). We find that the BERT base model outperforms simple versions of commonly used 
style measuring approaches like LIWC, punctuation frequency or character 3-grams. %
We invite the addition of complementary tasks and hope that this framework will facilitate the development of improved style-sensitive models and methods. 
Our data and code are available on GitHub.\footnote{\url{https://github.com/nlpsoc/STEL}}

\begin{table*}[th!]
\centering \small
\begin{tabular}{p{10mm} c | c p{25mm} p{24mm} p{27.9mm} p{26.8mm}}
\toprule
\textbf{Comp} & \textbf{Size} & \textbf{Order}  & \textbf{Anchor 1 (A1)} & \textbf{Anchor 2 (A2)} & \textbf{Sentence 1 (S1)}& \textbf{Sentence 2 (S2)} \\
\midrule
\textbf{formal/\-informal} & 815 & S1-S2 &  Are you one of their fans? & r u a fan of them or something? & Oh, and also, that young physician got an unflattering haircut. & Oh yea and that young dr got a bad haircut. \\
\midrule
\textbf{simple/\-complex} & 815 & S1-S2 & These rock formations are made of sandstone with layers of quartz. & These rock formations are characteristically composed of sandstone with layers of quartz. & The Odyssey is an ancient Greek epic poem attributed to Homer. & The Odyssey is an old Greek epic poem written by Homer. \\
\midrule
\midrule
\textbf{number substitution} & 100 & S2-S1 & $<$3 friends forever & $<$3 friends 4ever & D00d \$30 is heaps cheap, that must work out to just a couple of bucks an hour & Dude \$30 is heaps cheap, that must work out to just a couple of bucks an hour  \\
\midrule
\textbf{contrac\-tion} & 100 & S1-S2 & 
    In that time, it's become one of the world's most significant financial and cultural capital cities. %
    & In that time, it has become one of the world's most significant financial and cultural capital cities. %
    & Will doesn't refer to any particular desire, but rather to the mechanism for choosing from among one's desires. %
    & Will does not refer to any particular desire, but rather to the mechanism for choosing from among one's desires. \\ %
\bottomrule
\end{tabular}
\caption{\label{table:quad-examples}
\textbf{STEL Examples.} We give an example for each component (Comp) of STEL: Formal/informal and simple/complex for the more complex style dimensions as well as number substitution and contraction for the simpler style characteristics. The task is to order sentence 1 (S1) and sentence 2 (S2) to match the style order of anchor 1 (A1) and anchor 2 (A2). 
The correct order is given in the `Order' column. 
}
\end{table*}

\section{Related Work}

Linguistic style has been analyzed from different perspectives and along different dimensions. 
A speaker's style can, for example, be influenced by the situation, the speaker's choices, and/or the speaker's identity \cite{xu2015user, flekova-etal-2016-exploring,nguyen-etal-2016-survey,bell-audience-design}.
In NLP, previously analyzed style dimensions include formal/informal, simple/complex, abstract/concrete and polite/impolite \cite{pavlick-nenkova-2015-inducing, pavlick-tetreault-2016-empirical, paetzold-specia-2016-semeval, brooke-hirst-2013-hybrid, madaan-etal-2020-politeness}. 

Linguistic style is usually defined to be distinct from content. However, style is often found to be correlated with content (e.g., \newcite{gero-etal-2019-low}). To address this, some %
control for content with word-level %
paraphrases \cite{pavlick-nenkova-2015-inducing,xu2015user,niu-carpuat-2017-discovering}, topic labels (e.g., \newcite{AV_SimLearning_Attention}) or by avoiding the use of content-specific features (c.f.~\newcite{neal_aa-survey, masking-topic_AA}), others choose no or only limited control for content (e.g., \newcite{zangerle:2020, kang-hovy-2021-style}).
There has been considerable work in creating parallel datasets of (sentence level) paraphrases with shifting style, often using human annotations
\cite{xu-etal-2012-paraphrasing,  xu-etal-2016-optimizing, rao-tetreault-2018-dear, krishna-etal-2020-reformulating}. The task of generating paraphrases of text fragments with different style properties is sometimes also called \textit{style transfer}. %

There is little work on general evaluation benchmarks for style measuring methods. 
 \newcite{kang-hovy-2021-style} use style classification tasks to compare 5 language models. Only models that classify style into the given 15 dimensions can be evaluated.
 They do not control for content. %
Individually fine-tuned 
RoBERTa \cite{liu2019roberta} and BERT \cite{devlin-etal-2019-bert} classifiers for one style 
were outperformed by a fine-tuned T5 model \cite{2020t5} that was  jointly trained on multiple style labels. 
BERT/RoBERTa outperformed the T5 model on some styles (e.g., `sarcasm' and `methaphor').
Other related tasks are the PAN \textit{Authorship Verification} \cite{kestemont:2020} and \textit{Style Change Detection} \cite{zangerle:2020} tasks
which aim at identifying whether two documents or consecutive paragraphs %
have been written by the same author. In their current version both tasks do not control for topic. However, \newcite{kestemont:2020} controls for domain (here: `fandom' of the considered `fanfictions'). The best performing model for \newcite{kestemont:2020} was a neural LSTM-based siamese network \cite{boenninghoff:2020}, which is conceptually similar to some variants of sentence BERT \cite{reimers-gurevych-2019-sentence}.
The PAN setup assumes that authors tend to write in a relatively consistent style. Based on similar assumptions, the field of \textit{authorship attribution} wants to determine which author wrote a given document. %

Especially in authorship attribution, recurring style features include character n-grams, punctuation, average word length or function word frequency \cite{neal_aa-survey, 10.1093/llc/fqm020,10.5555/1527090.1527102}. %
Other recurring methods for style measurement include LIWC \cite{LIWC, Mark-My-Words, el-baff-etal-2020-analyzing}, and learned vector representations of words and sentences %
\cite{akama-etal-2018-unsupervised, stylometric_rep_AA, hay-etal-2020-representation}. \newcite{niu-carpuat-2017-discovering} suggests that style variations are already represented in commonly used neural embeddings.

Binary and more fine-grained style classification has been employed on word, text fragment as well as document level \cite{danescu-niculescu-mizil-etal-2013-computational, pavlick-nenkova-2015-inducing, xu2015user, pavlick-tetreault-2016-empirical, kang-etal-2019-male}. 
Traditionally, considered documents in authorship attribution were longer than 1,000 words (e.g., \newcite{eder_sizematters}), but recently there has been increased interest in text fragments with fewer than 300 words (e.g., \newcite{6705711, AV_SimLearning_Attention}).

\section{Style Evaluation Framework} 
\label{sec:stel-framework}
\label{sec:benchmark}
\label{sec:STEL-task}
We introduce the modular, 
fine-grained,
and content-controlled
similarity-based 
STyle EvaLuation framework (STEL). STEL tests a (language) model's ability to capture the style of a sentence. 

\paragraph*{{Modular Operationalization of Style.}}
\label{sec:stel-modular}
Style has previously been conceptualized in many different ways. From being defined as purely aesthetic in \newcite{biber_conrad_2009} 
to encompassing all forms of language variation, e.g., in \newcite{crystal_english_style}. 
We refrain from meddling in the style definition debate and instead use the broad notion of ``how vs. what'', i.e., how something is said as opposed to what is said. Inspired by \newcite{Campbell_featureClusters}, we use different \textit{characteristics} (i.e., more specific linguistic choices) as well as more general \textit{dimensions} of style (i.e., more complex combinations of style features).
By not only using complex style  dimensions, but also 
small scale and simpler  characteristics, STEL allows for very controlled and \textbf{fine-grained} testing. We can easily make sure that only the characteristics and no other aspects change (c.f.~Table \ref{table:quad-examples}).
Depending on one's goal and understanding of style, some \textit{components} (i.e., dimensions or characteristics) should be excluded and others should be added to this modular framework.
We exemplify the framework's more complex dimensions with the formal/informal distinction as this has been one of the most agreed upon dimensions of style \cite{Heylighen1999FormalityOL, labov_2006}. 
Additionally, we use the simple/complex dimension which has been used in connection to linguistic-stylistic choices as well \cite{jbp:/content/journals/10.1075/jaic.20014.haa, pavlick-nenkova-2015-inducing}. 
We exemplify the framework's simpler style dimensions (i.e., characteristics) with numb3r substitutions and contraction usage.  See Table \ref{table:quad-examples} for examples for each component. %

\paragraph*{Controlling for Content.} 
\label{sec:stel-content}
It is difficult to clearly separate style from content \cite{masking-topic_AA,gero-etal-2019-low}. 
Specific scenarios might correlate with both style and content. For example, in a job interview applicants might use a more formal style and talk more about their profession than in a more informal setting at a bar. Then, a 
model that generally rates texts about jobs as formal and texts about beverage choices as informal might perform well at style prediction.
In other words,
models that correctly use style features could sometimes be indistinguishable from those that use topical features. %
To control for content, we use parallel paraphrase datasets (\S \ref{sec:task-generation}), which consist of a set of sentences written in one style and a parallel set of sentences written in another.

\paragraph*{Task Setup.}
\label{sec:STEL-task-setup}
We test a method's style measuring capability with tasks of the setup shown in Figure \ref{fig:STEL-Task}. The sentences (S1 and S2) have to be ordered to match the order of the  anchor sentences (A1 and A2). Here, `r u' (A1) and `Oh yea' (S2) are written in a more informal style than their respective paraphrases A2 (`Are you') and S1 (`Oh, and also'). Thus, the correct order is S2, then S1. We call this setup the \textit{quadruple setup}. Additionally, we explore a second task setup, the \textit{triple setup}, which leaves out anchor 2 (A2). There, the task is to  decide which of the two sentences matches the style of anchor 1 (A1) the most. The two different setups are similar to the triple and quadruple training instances in the field of {metric learning}, e.g., \cite{JMLR:v21:19-678, law:hal-01346190, kaya2019deep}.

\section{Task Generation} %
\label{sec:task-filtering}

We describe the task instances of STEL: First, we generate potential task instances (\S \ref{sec:task-generation}). Second, we describe problems with the generated instances (i.e., ambiguity in \S \ref{sec:generated-ambiguity}). Third, we filter out the problematic instances via crowd-sourcing (\S \ref{sec:annotation-setup}). %

\subsection{Potential Task Instances}
\label{sec:task-generation}
We generate potential task instances on the basis of parallel paraphrase datasets written in style 1 and style 2 respectively.  For each style 1/style 2 paraphrase pair (anchors in Table \ref{table:quad-examples}), we randomly select another sentence pair (sentences in Table \ref{table:quad-examples}). Again randomly, we decide which of the anchor pair is anchor 1 (A1) and which is anchor 2 (A2) and fix that ordering for all future considerations. We do the same for the sentence pair. The answer to the STEL task (Figure \ref{fig:STEL-Task}) is labeled as S1-S2 if A1 was taken from the same style set as S1, e.g., both from style 1. Otherwise the order is reversed.

\paragraph*{Formal/Informal Dimension.} %
We use the test and tune split of the Entertainment\_Music GYAFC subcorpus \cite{rao-tetreault-2018-dear} as the parallel paraphrase dataset. 
It consists of a set of informally phrased sentences and a parallel set of crowd-sourced formal paraphrases. 
We generate 918 potential STEL formal/informal task instances.  %

\paragraph*{Simple/Complex Dimension.} We use the test and tune split from \newcite{xu-etal-2016-optimizing}.
It consists of English Wikipedia sentences and 8  crowd-sourced simplifications per sentence. For each Wikipedia sentence, we randomly draw the parallel paraphrase out of the 8 simplifications. We discard sentences that are too close to the original via the character edit distance of 3 or lower. From this parallel paraphrase dataset, we generate 1195 potential STEL simple/complex task instances.

\paragraph*{Contraction Characteristic.}
We generated the parallel contraction dataset from the December 2018 abstract dump of English Wikipedia\footnote{\url{https://archive.org/details/enwiki-20181220}}. The Wikipedia style guide discourages contraction usage and provides a dictionary with contractions that should be avoided.\footnote{\url{https://en.wikipedia.org/wiki/Wikipedia:List_of_English_contractions}} We use an adapted version\footnote{See the Appendix for the filtered contraction list.
} to select 100 sentences where an apostrophe is present and a contraction is possible.
Such a sentence could be ``It is near Thomas's car''.
For each sentence, we generate a parallel sentence with a contraction, e.g., ``It's near Thomas's car'', c.f.~Table \ref{table:quad-examples}.

\paragraph*{Number Substitution Characteristic.}
The character by number substitution task instances were semi-automatically generated from the Reddit comment corpus of the months May 2007-September 2007, June 2012, June 2016 and June 2017 taken from the Pushshift dataset \cite{Baumgartner_Zannettou_Keegan_Squire_Blackburn_2020}. We selected a pool of potential sentences where words contained character substitution symbols ({4,3,1,!,0,7,5}) or  are part of a manually selected list of number substitution words (see  Appendix). Then, we manually filtered out sentences without number substitutions (e.g., common measuring units or product numbers).  We selected 100 sentences, 50 of which were selected to contain at least one additional number that is not part of a number substitution word (e.g., Anchor 1 in Table \ref{table:quad-examples}). This setup ensures that the task is not as simple as checking whether there are numbers present in the sentence. To generate the parallel phrases, we manually translated the sentences to contain no number substitutions.
As we looked for naturally occurring number substitution words, we decided to keep word pairs that contain additional changes besides number substitution. For example,
generally different spelling (e.g., `d00d', `dude') or phonetic spelling (e.g., `str8', `straight'). We decided to replace the number substitution symbols with characters only --- e.g., not with punctuation marks as seen in `s1de!!!!!1!' -> `side!!!!!1!'.

\begin{figure}[t]
    \centering
    \includegraphics[width=0.49\textwidth]{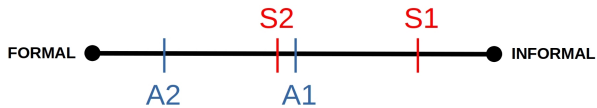}
    \caption{\textbf{Triple Problem.} Tasks are generated from sentence pairs (A1, A2) and (S1, S2) that are split along the same style dimension (e.g., formal/informal). For each pair, only the order on the axis (e.g., S2 $<$ S1) but not the absolute localization is known. This might lead to a wrong generated label for the triple setup. Here, removing A2 leads to S2 being stylistically closer to A1, whereas the generated label would be S1.
    }
    \label{fig:triplet_problem}
\end{figure}

\subsection{Ambiguity}
\label{sec:generated-ambiguity}
Manual inspection shows that the generated potential task instances of the formal/informal and simple/complex dimension contain ambiguities: 
(i) Some are a result of unclear or very fine
distinctions between the two parallel styles in the original data. For example, \textit{``Each band member chose an individual number as their alias towards the end of 1997.''} and \textit{``Towards the end of 1997, each band member chose an individual number as their alias.''}. 
The first is labelled as written in a simpler style in \newcite{xu-etal-2016-optimizing}. However, putting ``towards the end of 1997'' at the beginning of the sentence could also be understood as structuring the sentence more clearly, and thus as simpler. After manual inspection, ambiguities 
seem to be more prevalent for the simple/complex than the formal/informal dimension. %
(ii) Other ambiguities are the result of entangled additional linguistic  components. For example, consider the potential task instance (A1) \textit{``He's supposed to be in jail!''}, (A2) \textit{``I understood he was still supposed to be incarcerated.''} and (S1) \textit{``green day is the best i think''}, (S2) \textit{``I think Green Day is the best.''}. The sentences are clearly split along the formal/informal dimension leading to the label S1-S2. Still, (A1) and (S2) could also be understood as being written in a more decisive tone than (A2) and (S1) leading to the order S2-S1. 
\label{sec:triple-problem}

We find that the triple setup has additional theoretical limitations that can lead to ambiguity: 
Consider the `Triple Problem' in Figure \ref{fig:triplet_problem} where S2 is labelled as formal and A1 is labelled as informal. Removing A2, to get from the quadruple to a triple setup, will leave A1 closer to S2, contrary to the original labelling (see also the previous example in  (ii)). %
Additionally, having fewer sentences in the triple setup increases the chance of a random correlation with a  different linguistic component (similar to the `decisive tone' in example (ii)).

\begin{table*}[th]
    \centering \small
    \subfloat[Results on the sample and total of task instances \label{table:results_annotator-agreement}
    \label{table:results-annotation}]{
     \begin{tabular}{l || c c c c | c || c c | c } 
     \toprule
     & \multicolumn{5}{c||}{\textbf{Sample}} & \multicolumn{3}{c}{\textbf{Total}} \\
     & \multicolumn{5}{c||}{\textbf{}} & \multicolumn{3}{c}{\textbf{}}  \\
      & \multicolumn{2}{c}{Triple} & \multicolumn{2}{c|}{Quadruple}  & n & \multicolumn{2}{c|}{Quadruple} & n\\ %
     \midrule
     Dim & {$\kappa$} & acc. & {$\kappa$} & acc. &  & {$\kappa$} & acc. &\\
     \midrule %
     all & {0.29} & {0.62} & {0.35} & {0.78} & 602 & {0.30} & {0.77} & 2113\\
     \hspace{2pt} c & 0.19 & 0.51 & 0.16 & 0.68 & 301 & 0.17 & 0.68 & \hspace{4pt}918 \\
     \hspace{2pt} f & 0.39 & 0.74 & 0.51 & 0.89 & 301  & 0.48 & 0.90 & 1195 \\
     \bottomrule    
    \end{tabular}
    }
    \qquad
    \subfloat[Subsample analysis\label{table:error-analysis}]{ \small %
    \begin{tabular}{c c c c}
    \toprule
    Triple & Quadruple & Dimension & Share  \\
    \midrule
    \multirow{2}{*}{\xmark} & \multirow{2}{*}{\cmark}
      & formal &  $0.196$  \\ %
     & & complex &  $0.312$  \\ %
    \multirow{2}{*}{\cmark} & \multirow{2}{*}{\xmark} & formal &  $0.047$  \\ %
     &  & complex &  $0.140$  \\ %
    \multirow{2}{*}{\xmark} & \multirow{2}{*}{\xmark} & formal & $0.066$ \\ %
     &  & complex &  $0.179$  \\ %
    \multirow{2}{*}{\cmark} & \multirow{2}{*}{\cmark} & formal &  $\boldsymbol{0.691}$ \\ %
     &  & complex &  $\boldsymbol{0.369}$  \\ %
    \bottomrule
    \end{tabular}} %
    \caption{\textbf{Annotation Results.} 
    We filter out ambiguous task instances via annotations.
    In (a), we display inter-annotator agreement (Fleiss's $\kappa$) and annotation accuracy (acc.) for the sample 
    and total of potential task instances on the quadruple and triple setup for the simple/complex (c) and the formal/informal (f) dimensions. We also display the number of task instances per dimension (n).
    In (b), we display the share of all combinations of correct (\cmark) and wrong (\xmark) annotations per dimension and task setup. 
    The union of \cmark \cmark and \xmark \cmark cases make up a majority.  
    }
\end{table*}

\subsection{Removing Ambiguity}
\label{sec:annotation-setup}

Using crowd-sourced annotations, we filter the previously discussed ambiguity out of the potential formal/informal and simple/complex task instances. 
The simpler STEL tasks (contraction and number substitution) mostly differ in the amount of apostrophes and numbers (e.g., Table \ref{table:quad-examples}).
As a result, we expect the style characteristics to contain little to no ambiguity and do not filter those further.

\paragraph*{Annotation Tasks.}
For both the triple and quadruple setup we collected annotations on a subsample of all generated task instances (301 simple/complex and formal/informal instances respectively). 
Then, %
we annotated a larger set of task instances on the quadruple setup alone. 
Based on performance results from the subsample (`Annotation Results'), we had 617 and 894 more task instances annotated for the formal/informal and simple/complex dimension respectively. %

\paragraph*{Annotation Setup.} We used annotations from 839 different \textit{Prolific}\footnote{\url{https://www.prolific.co/}} crowdworkers with 5 distinct annotators per potential task instance. We paid participants 10.21\pounds/h\footnote{above UK minimum wage of 8.91\pounds/h at the time of the study (April 2021), see \url{https://www.gov.uk/national-minimum-wage-rates}} on average. All annotators were native English speakers as we assume them to have a better intuition about their language. See the Appendix for further detail.

\label{sec:annotation-results}

\paragraph*{Annotator Agreement.} In Table \ref{table:results_annotator-agreement}, we report inter-annotator agreement with Fleiss's $\kappa$ \cite{fleiss1971measuring} as $\kappa$ allows %
different items to be rated by different sets of raters. 
Inter-annotator agreement is only moderate. This does not mean that the annotations are of poor quality. As discussed in \S \ref{sec:generated-ambiguity}, our generated data contains ambiguous, noisy or faulty task instances.
 Manual inspection confirms that low annotator agreement is a sign of ambiguity (see also `Annotation Analysis' Table in Appendix).
This problem is more pronounced for the simple/complex than the formal/informal dimension. We ensured annotator quality with screening questions (appendix Table 1) and by selecting annotators with the highest platform-internal rating.

\paragraph*{Annotation Results.} Results are reported in Table \ref{table:results-annotation}. Annotation accuracy is the share of correctly annotated task instances (by a majority of at least 3) out of all potential task instances. 

The accuracy and the inter-annotator agreement are considerably higher for the formal/informal dimension (Table \ref{table:results-annotation}) than for the simple/complex dimension. 
This aligns with our expectation of more ambiguity in the simple/complex task instances (c.f.~\S \ref{sec:generated-ambiguity}(i)). %
Similarly, our expectations regarding theoretical problems with the triple setup (\S \ref{sec:triple-problem}) are  confirmed: Accuracy for the sample is generally higher for the quadruple than the triple setting. %
There are more examples where the quadruple setup was correctly annotated but the triple setup was not (\xmark\cmark in Table \ref{table:error-analysis}), than there are for the opposite kind (\cmark\xmark).

As a consequence, the annotation of the bigger set of task instances was only done on the quadruple setup. On the total set of potential task instances (which includes the sample) we obtained similar accuracy and annotator agreement as on the sample (see Table \ref{table:results-annotation}). 
We filter the potential task instances by only keeping those that were correctly  annotated by a majority (i.e., at least 3/5). This leaves 822  task instances for the formal/informal and 815 for the simple/complex dimension. We randomly remove 7 task instances from the formal/informal dimension for equal representation of the two style dimensions.
In the following, and under the name STEL, we will only consider the quadruple setup on the 1830 filtered task instances (i.e., 815, 815, 100 and 100 for simple/complex, formal/informal, number substitution and contraction respectively). %

\section{Evaluation}
\label{sec:ex-setup}

\begin{table*}[th]
    \small
    \centering
    \begin{tabular}{ l|l | l l | l l | l | l | l l  }
     \toprule
       & all & \multicolumn{2}{c|}{formal} & \multicolumn{2}{c|}{complex} & nb3r & c'tion & \multicolumn{2}{c}{random} \\
       & & filter & full & filter & full &  &  & filter & full \\
     \midrule
     BERT uncased & \textbf{0.74} & \textbf{0.79} & 0.77          & \textbf{0.65} & 0.63  & {0.90} & 0.90 & \textbf{0} & 0\\ 
     BERT cased     & \textbf{0.77} & \textbf{0.82} & 0.81  & \textbf{0.68} & 0.64 & \textbf{0.92} & \textbf{1.0} & \textbf{0} & 0\\
     RoBERTa        & 0.61 & 0.63 & 0.62   & 0.54 & 0.53 & 0.62 & 0.98 & \textbf{0} & 0\\
     SBERT mpnet & 0.61 & 0.64 & 0.62 & 0.53 &  0.52 & 0.71 & 0.84 & \textbf{0} & 0 \\
     SBERT para-mpnet & 0.68 & 0.73 & 0.72 & 0.55 & 0.54 & \textbf{0.95} & \textbf{1.0} & \textbf{0} & 0 \\
     USE            & 0.59 & 0.59 & 0.58   & 0.55 & 0.52 & 0.58 & 0.85 & \textbf{0.00} & 0.00 \\  %
     \midrule     
     BERT uncased NSP      & 0.66 & 0.72 & 0.71   & 0.59 & 0.57  & 0.67 & 0.70 & 0.10 & 0.12 \\  %
     BERT cased NSP   & 0.71 & \textbf{0.79} & 0.77 & 0.60 & 0.58  & 0.77 & {0.96}  & 0.02 & 0.02 \\  %
     \midrule
     char 3-gram    & 0.55 & 0.58 & 0.57          & 0.52 & 0.50 & 0.50 & 0.64 & 0.05 & 0.05 \\ %
     word length    & 0.58 & 0.53 & 0.53          & 0.59 & 0.57 & 0.50 & 0.94 & 0.08 & 0.08 \\ %
     punctuation    & 0.56 & 0.58 & 0.58   & 0.50 & 0.49 & 0.50 & 0.92 & 0.38 & 0.39 \\  %
     \midrule
     LIWC           & 0.55 & 0.52 & 0.52          & 0.52 & 0.52 & 0.50 & \textbf{0.99} & 0.09 & 0.09 \\ %
     LIWC (style)   & 0.50 & 0.52 & 0.52          & 0.50 & 0.50 & 0.50 & 0.50 & 0.62 & 0.64 \\ %
     LIWC (function) & 0.53 & 0.48 & 0.48    & 0.52 & 0.51 & 0.50 & \textbf{1.0} & 0.28 & 0.28 \\ %
     deepstyle      & 0.66 & 0.71 & 0.70           & 0.55 & 0.52  & 0.84   & 0.96 & \textbf{0} & 0\\     
     \midrule
     POS Tag        & 0.52 & 0.53 & 0.53          & 0.52 & 0.52 & 0.50 & 0.50 & 0.20 & 0.20 \\ %
     share cased    & 0.56 & 0.55 & 0.54         & 0.53 & 0.51 & 0.50  & \textbf{1.0} & 0.08 & 0.08 \\ %
     edit dist      & 0.54 & 0.56 & 0.56          & 0.52 & 0.51 & 0.50 & 0.39 & 0.08 & 0.07 \\ %
     \bottomrule
    \end{tabular}
    \caption{\textbf{STEL Results.} We display STEL accuracy 
    for different language models and methods. %
    Random performance is at 0.5.
    The share of task instances for which a method decides randomly as it can not decide between the two options (`=' in Equation \ref{eq:sim-vector-length}) is given in the `random' column. %
    Both the performance on the set of task instances before (full) and after crowd-sourced filtering (filter) is displayed. The two best accuracies are boldfaced.
     The BERT-based models perform the best, followed by the ``deepstyle'' style sentence embedding method. On average, methods perform best for the c'tion and worst for the simple/complex dimension.  %
     }
    \label{table:results-stle-models}
\end{table*}

We use our STEL framework to test several models and methods that could be expected to capture style information (\S  \ref{sec:style-measures}). We describe how the models decide the STEL tasks (\S \ref{sec:decision-alternatives}) and discuss their performance on STEL (\S \ref{sec:model-results}).  

\subsection{Style Measuring Methods}
\label{sec:style-measures}
We describe methods and models that can be used to calculate a (style) similarity. %
Given two sentences, the methods return a similarity  value between 0 and 1 or -1 and 1 (when using cosine similarity), where 1 represents the highest similarity. 

\paragraph*{Language Models.} We use the base \textit{BERT uncased} and base \textit{BERT cased} model \cite{devlin-etal-2019-bert}. We calculate the mean over the subwords in the last hidden layer to generate two sentence embeddings. Then, we use cosine similarity to compare the sentences. 
We do the same with the cased \textit{RoBERTa} base model \cite{liu2019roberta}. Additionally, we compare to the sentence BERT  
`all-mpnet-base-v2' (\textit{SBERT mpnet})\footnote{best performing pre-trained sentence embedding in September 2021, see \url{https://www.sbert.net}} and 
`paraphrase-multilingual-mpnet-base-v2' (\textit{SBERT para-mpnet})\footnote{best performing embedding trained on paraphrase data}
models \cite{reimers-gurevych-2019-sentence,reimers-gurevych-2020-making}. Like BERT, MPNet uses a transformer architecture, but with a permuted instead of a masked language modeling pre-training task \cite{mpnet}. Furthermore, we experiment with the universal sentence encoder (\textit{USE}) from  \newcite{cer-etal-2018-universal}. 

\paragraph*{Authorship Attribution Methods.} The following methods are inspired by successful or commonly used approaches in authorship attribution \cite{neal_aa-survey, sari-etal-2018-topic}. We use character 3-gram similarity by calculating the cosine similarity between the frequencies of all \textit{character 3-gram}s. We calculate the \textit{word length} similarity via the average word lengths $a$ and $b$ of two sentences: $1-|a - b|/\text{max}(a,b)$.
We calculate the \textit{punctuation} similarity by using the cosine similarity between the frequencies of punctuation marks \{',:,`,',\_,!,?,;,.,",(,),-\}, taken from \newcite{sari-etal-2018-topic}. 

\paragraph*{LIWC-based Style Methods.} %
LIWC categories have previously been used as style features \cite{doi:10.1177/026192702237953}.
We use LIWC 2015 \cite{LIWC} for (a) \textit{LIWC} similarity by taking the cosine similarity between the complete LIWC frequency vectors, (b) \textit{LIWC (style)} similarity by taking the cosine similarity between the 8 dimensional binary LIWC style vectors (1 if a word of the category is present in the sentence, 0 otherwise) proposed in \newcite{danescu_echoes-power}, (c) \textit{LIWC (function)} similarity by taking $1 -$ the difference between the relative frequencies of function words. 
Function words have previously been used as a proxy for style  \cite{neal_aa-survey}. 

\paragraph*{Other Methods.} 
We also experiment with the ``\textit{deepstyle}'' model \cite{hay-etal-2020-representation} by taking the cosine similarity between the style vector representations.
Additionally, we consider the following sentence features: NLTK \textit{POS Tags} \cite{nltk} and \textit{share of cased} characters (e.g., \newcite{sari-etal-2018-topic}) via the cosine similarity between the frequency vectors and 1 - the difference between the proportion of cased characters respectively. 
We also include the \textit{edit dist}ance as a simple baseline.

\subsection{Similarity-based Decision} %
\label{sec:decision-alternatives} 

To determine an answer for a STEL task in the quadruple setup, the methods need to order two sentences (Figure \ref{fig:STEL-Task}). We do this by calculating the similarities (sim) between Anchor 1 ($\text{A}1$), Anchor 2 ($\text{A}2$), Sentence 1 ($\text{S}1$) and Sentence 2 ($\text{S}2$). We decide for the order $\text{S}1$-$\text{S}2$ if 
\begin{equation}
    \small
    \begin{aligned}
     \label{eq:sim-vector-length}
        \small
        &(1-\text{sim}(\text{A1}, \text{S1}))^2+(1-\text{sim}(\text{A2}, \text{S2}))^2 < \\
        & (1-\text{sim}(\text{A1}, \text{S2}))^2+(1-\text{sim}(\text{A2}, \text{S1}))^2
    \end{aligned}
\end{equation}
For the `$>$' case we use the order  $\text{S}2$-$\text{S}1$, for `=' ordering is settled randomly (c.f., `random' in Table \ref{table:results-stle-models}).
See the Appendix %
for a proof sketch after transforming similarities to distances.

\subsection{Results}
\label{sec:model-results}

Performance results are shown in Table \ref{table:results-stle-models}. The accuracy is a weighted mean of 0.5 (proportional to the share of undecided instances, c.f. ‘random’ in Table 3) and the accuracy in the decided cases. Random guessing would show an accuracy of 0.5 exactly. Stylistic differences can be subtle for the STEL dimensions and we expect this to be a hard task to solve. In contrast, the STEL characteristics (i.e., contraction and number substitution) should be easier to solve (via detecting an additional apostrophe or number) and are especially interesting for model error analysis.
Note: We do not make general quality judgements because models were not trained on the components of STEL and were often not even meant to measure style directly. 

\paragraph*{The BERT base model encodes style information.} The best performing cased BERT base model has an overall accuracy of 0.77.  RoBERTa, a successor of BERT, includes BERT's training data \cite{liu2019roberta}. 
However, the cased RoBERTa base model does less well (0.61, $p<0.001$ with McNemar's test \cite{mcnemar}). 
A possible explanation of RoBERTa's reduced performance might be the removal of the next sentence prediction (NSP) task. Closer sentences could generally be more similar in style than a different random sentence --- possibly making the NSP a valuable learning objective for style similarity learning. To further look at this, we experiment with the \textit{BERT NSP} head on the cased and uncased base model. 
For the quadruple setup, we calculate the four `similarity' values as described in Equation \ref{eq:sim-vector-length} by using the predicted softmax probability that A1 is followed by S1 for sim(A1, S1). The other similarities are calculated equivalently.
Interestingly, BERT's cased NSP head (accuracy of 0.71) performs better than  RoBERTa ($p<0.001$) across the STEL tasks. The effect of training objectives on learning style information could be explored in future work.

\paragraph*{(Semantic) Sentence Embedding Methods perform well.} SBERT para-mpnet (0.68) trained on the paraphrase data performs better than SBERT mpnet (0.61, $p<0.001$) and USE (0.59, $p<0.001$). Overall, SBERT para-mpnet is the third best performing model after the base BERT models and the best performing model in the nb3r dimension.  In future work, it could be interesting to explore the effect of different training data on the performance of embedding models.

\paragraph*{LIWC alone does not perform well.} 
On the style dimensions LIWC performs similar to the random baseline. Possibly because the LIWC methods often find no difference between the two possible orderings (10\%, 71\% and 32\% of tasks). The difference between the three LIWC-based methods is not significant ($p>0.05$). 
Future work could explore models that consider more fine-grained differences between LIWC categories. 

\paragraph*{Authorship attribution methods perform better than random.} 
Character 3-grams and punctuation perform at 0.58 accuracy on the formal/informal dimension. Considering some of the informal examples, punctuation seems to be one of the most prominent visible changes from a formal to an informal style (see Appendix). %
Interestingly, word length is the method that most clearly performs better on the simple/complex than the formal/informal dimension. %
This aligns with the intuition that shorter words are a sign of a simpler style as found in \newcite{paetzold-specia-2016-semeval}. 

\paragraph*{Casing encodes style information.}
The uncased performs worse than the cased BERT model (0.74 vs. 0.77, p=0.008). %
Additionally, the cased letter ratio performs slightly better than random for the formal/informal dimension (0.55) and perfect for the contraction characteristic (1.0): When the sentence consists of fewer lower cased characters (as a result of removing them when using contractions), the share of upper cased characters increases. %

\paragraph*{Style embedding yields promising results.} The method ``deepstyle'' \cite{hay-etal-2020-representation} performs well across STEL components (0.66). It performs the worst on the simple/complex dimension (0.55). The method embeds sentences in a vector space where texts by ``similar'' authors are similarly embedded. %
In the training data (blog and news articles), authors might not consistently use one style over the other. %
The difference between same author and same style could be explored in future work.

\paragraph*{Less ambiguous task instances reach higher accuracy values.} Table \ref{table:results-stle-models} (c.f.~`full') shows the accuracy of the style measuring methods for the complete set of potential task instances before filtering out ambiguity (\S \ref{sec:task-generation}). The accuracies are the same or lower  than the crowd-validated task instances in STEL. %
The differences are more pronounced for the simple/complex than the formal/informal dimension. This aligns with the higher (expected) ambiguity in the simple/complex dimension (\S \ref{sec:generated-ambiguity} and \S \ref{sec:annotation-results}).  In general, we recommend to use the filtered STEL task with less ambiguity for testing. 

\section{Limitations and Future Work}

Our illustrative set of task instances does not cover all possibilities of style variation. Future work could extend STEL to cover additional style dimensions or more fine-grained task instances using several sources of data.

The STEL task instances for one style component can  contain correlations with unconsidered (style) components.
Consider the following task instance (shortened for readability):
(A1) \textit{\say{Forty-nine species of pipefish [...] have been recorded.}},
(A2) \textit{\say{Forty-nine type of pipefish [...] have been found}}, 
(S1) \textit{\say{Patients [...] must have their liver checked for damage and other side effects.}} and (S2)
\textit{\say{[...] patients [...] must be monitored for liver damage and other possible side effects.}}. 
(A2) and (S1) are the simpler version of (A1) and (S2) \cite{xu-etal-2016-optimizing}. Additionally, the sentences vary along other aspects: (A2) is missing the punctuation mark and includes a misspelling. (S1) is different in content from (S2) as (S1) is only considering effects on the liver while (S2) also includes other side effects. However, those aspects did not change the label given by the annotators (S2-S1) and should mostly be secondary to the considered style dimension.

With STEL, language models and methods 
are tested only on whether they capture clear differences in style when content is approximately the same. When there are also content differences, such models might put more emphasis on content than stylistic aspects. %
Our framework could be extended to allow testing for whether a model prefers style over content (e.g., with a new task format  
where sentence 1 is closer in content to anchor 1 but closer in style to anchor 2, c.f.~Figure \ref{fig:STEL-Task}). 

STEL could also be extended to test for individual author styles and style variation related to the social or regional background of authors (e.g., different age groups). For example, %
by including sentence pairs with the same content but written by different authors. 
Current and future dimensions could also be extended by a train/dev/test split to enable training on the task directly. Further, STEL could be enriched by including longer texts (e.g., paragraphs or documents) as anchor and alternative sentences.

\section{Conclusion}

Style is an integral part of language. However, there are only few benchmarks for linguistic style. In this work, we introduce STEL, a modular, content controlled and fine-grained similarity-based style evaluation framework. 
Out of the evaluated language models and methods, the cased BERT base model performs the best on STEL. Simpler sentence features perform close to the random baseline. 
STEL includes two general style dimensions  and two specific style characteristics. We hope that this framework will grow to include an even  more exhaustive representation of linguistic style and will facilitate the development of improved style(-sensitive) measures. %

\subsection*{Task Usage}

When using this task, please also cite the original datasets the tasks were generated from: (1) \newcite{rao-tetreault-2018-dear} for the formal/informal component and (2) \newcite{xu-etal-2016-optimizing} for the simple/complex component. (1) also needs the permission for usage of the ``L6 - Yahoo! Answers Comprehensive Questions and Answers version 1.0 (multi part)''\footnote{\url{https://webscope.sandbox.yahoo.com/catalog.php?datatype=l}}.

\section*{Ethical Considerations}

The STEL tasks are based on datasets \cite{rao-tetreault-2018-dear, Baumgartner_Zannettou_Keegan_Squire_Blackburn_2020, xu-etal-2016-optimizing} from popular online forums and web pages (Yahoo! Answers, Reddit, Wikipedia). However, the user demographics on these platforms are often skewed towards particular demographics. For example, Reddit users are more likely to be young and male.\footnote{\small {https://www.journalism.org/2016/02/25/reddit-news-users-more-likely-to-be-male-young-and-digital-in-their-news-preferences/}}  
Thus, our dataset might not be representative of (English) language use across different social groups.
Further, the usage of posts from online platforms without explicit consent from users might lead to (among others) privacy concerns. 
    The Wikipedia simplifications and formal Yahoo! Answers paraphrases were generated by consenting crowdworkers \cite{xu-etal-2016-optimizing, rao-tetreault-2018-dear}. We expect the sentences that were extracted from Wikipedia for the contraction dimension and for the complex/simple dimension to lead to minimal privacy concerns as they were meant to be read and copied by a broader public.\footnote{\url{https://en.wikipedia.org/wiki/Wikipedia:Copyrights}} \newcite{rao-tetreault-2018-dear} and the nb3r dimension do not include user names. However, we acknowledge that users might be identifiable from the exact wording of posts. We removed nb3r substitution instances that included Reddit user names. 
    We hope the ethical impact of reusing the already published  \newcite{rao-tetreault-2018-dear} dataset to be small.

\section*{Acknowledgements}
We thank the anonymous EMNLP reviewers for their  helpful feedback. We thank Yupei Du and Qixiang Fang for the productive discussions and their equally helpful feedback. This research  was supported by the ``Digital Society - The Informed Citizen'' research programme, which is (partly) financed by the Dutch Research Council (NWO), project 410.19.007. Dong Nguyen was supported by the research programme Veni with
project number VI.Veni.192.130, which is (partly) financed by the Dutch Research Council (NWO).

\bibliography{anthology,custom}
\bibliographystyle{acl_natbib}

\appendix

\clearpage

\section{Removing Ambiguity}
\begin{figure}[th]
    \centering
    \includegraphics[width=\columnwidth]{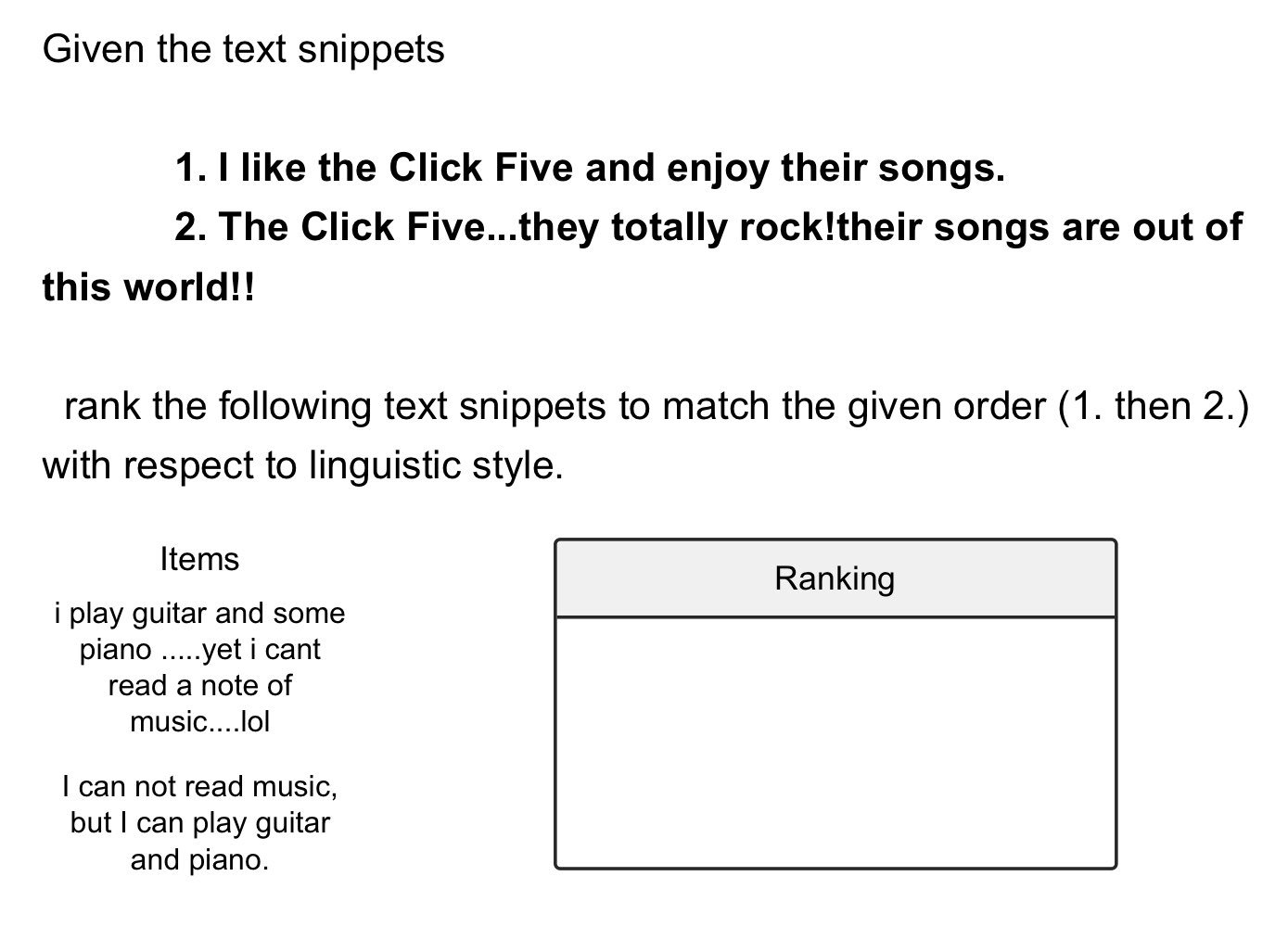}
    \caption{\textbf{Example survey question for the quadruple setup.} This is an example of what was shown to the crowdworkers.}
    \label{fig:quad_question}
\end{figure}

\begin{figure}[th]
    \centering
    \includegraphics[width=\columnwidth]{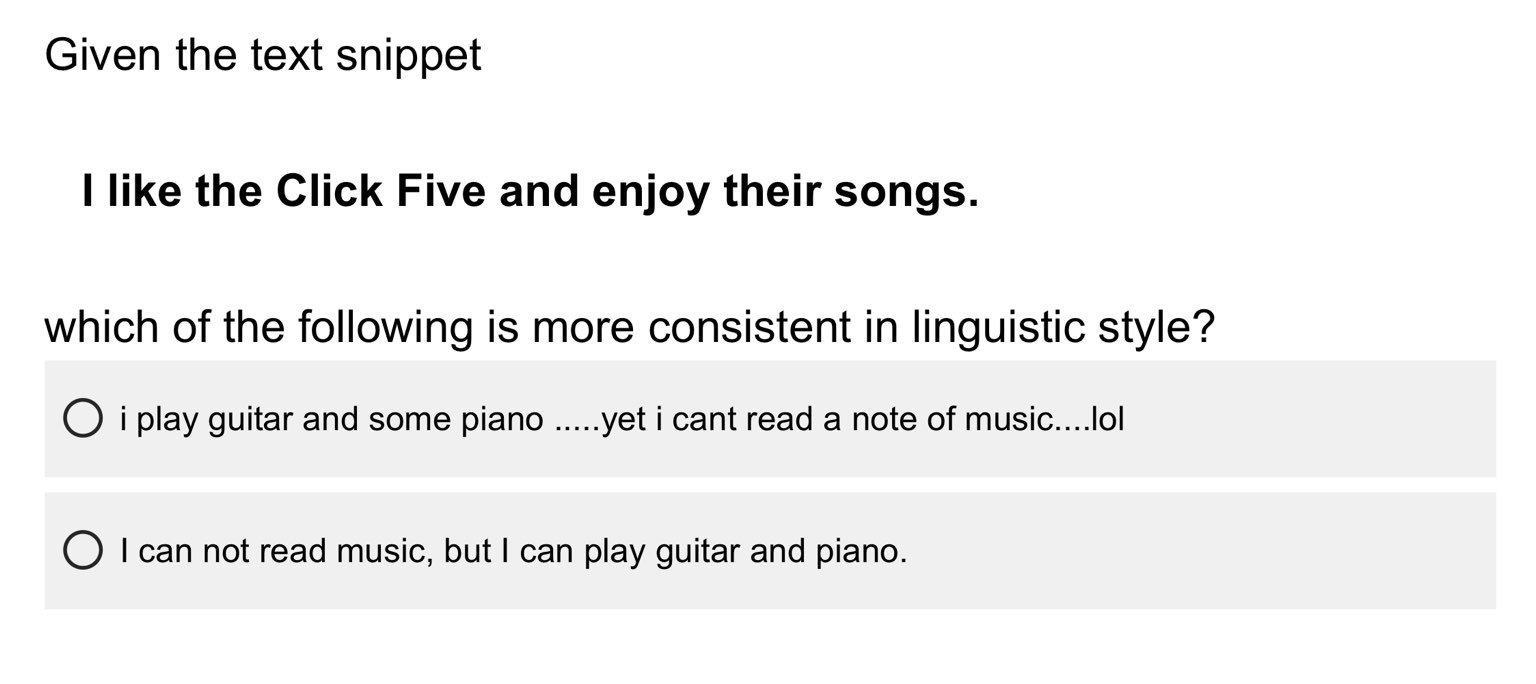}
    \caption{\textbf{Example survey question for the triple setup.} This is an example of what was shown to the crowdworkers.}
    \label{fig:trip_question}
\end{figure}

\begin{figure}
    \centering
    \includegraphics[width=\columnwidth]{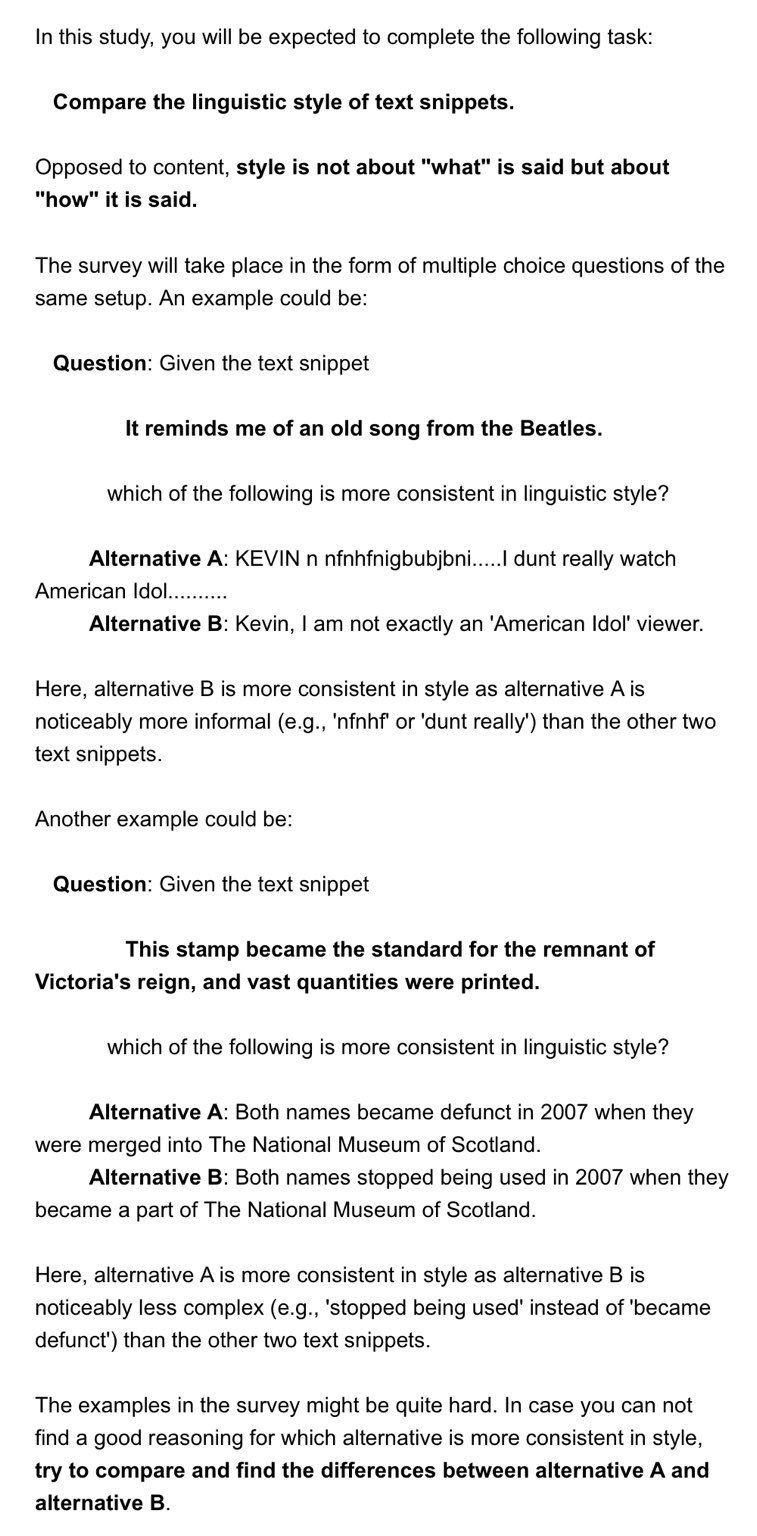}
    \caption{\textbf{Survey task description for the triple setup.} This is a copy of what was shown to the crowdworkers.}
    \label{fig:trip_description}
\end{figure}

\begin{figure}
    \centering
    \includegraphics[width=\columnwidth]{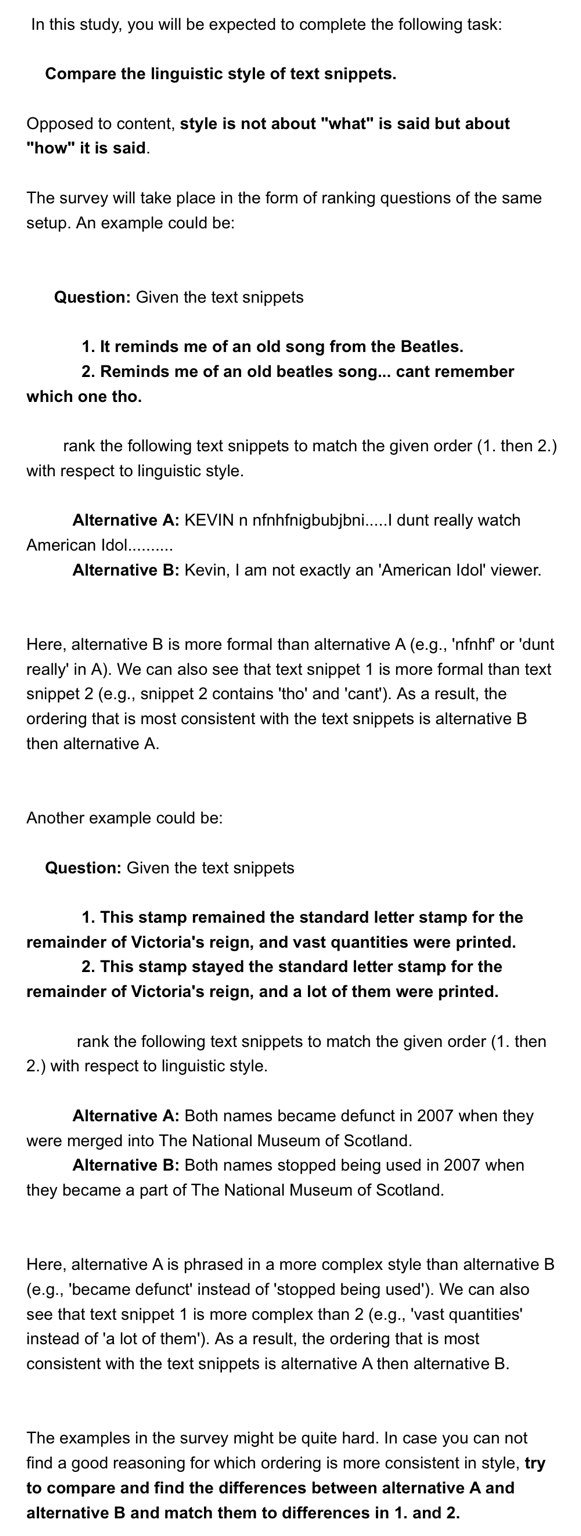}
    \caption{\textbf{Survey task description for the quadruple setup.} This is a copy of what was shown to the crowdworkers.}
    \label{fig:quad_description}
\end{figure}

\paragraph*{Annotation Setup Information.}
Prolific\footnote{\url{https://www.prolific.co/}} crowdworkers could participate up to 5 times in annotating different generated tasks from the formal/informal and simple/complex style dimensions. Each time a participant was asked to annotate 14 potential task instances as well as 2 additional screening questions. The screening questions were randomly sampled from a list of 10 screening questions (see Table \ref{table:screen_questions}). The screening questions were manually created and then unanimously and correctly answered by 3 lab-internal annotators in the triple setting.  We filtered out all crowdworkers that wrongly answered any of the screening questions. We display the task description (Figure \ref{fig:trip_description} and \ref{fig:quad_description}) as well as the phrasing of the questions (Figure \ref{fig:trip_question} and \ref{fig:quad_question}). Participants were payed 10,21£/hour on average (above 8.91£ UK minimum wage) and gave consent to the publication of their annotations. We required annotators to be native speakers as we assume them to have a better intuition about their language than non-native speakers: %
During study design, we conducted a pilot study with 8 different non-native annotators. Several felt their English-speaking abilities were insufficient for the task. The study projected a higher perceived and measured difficulty of the simple/complex dimension. As a result we required annotators to be native speakers and generated more potential simple/complex than formal/informal tasks.

\begin{table}[bh]
    \centering \small
     \begin{tabular}{l c c c } 
     \toprule
     & \multicolumn{2}{c}{Sample Acc.} \\
     \midrule
     Dim & Triple & Quadruple \\ 
     \midrule
      all &  \textbf{0.68} (+.053) & \textbf{0.85} (+.067) \\
       \hspace{2pt} complex &   0.54 (+.033) &   0.73 (+.044)\\
       \hspace{2pt} formal &  0.78 (+.042) &  0.94  (+.050)\\
     \bottomrule
    \end{tabular}%
\caption{\label{fig:opposite-filter} \textbf{Subsample Accuracy with Opposite Filtering.} The table displays the accuracy results on the STEL tasks that were correctly annotated in the opposite setup, corresponding to 472 and 375 task instances for the triple and quadruple setup respectively. We display the increase over the accuracy on the full sample in brackets (c.f.~`Sample' results in Table \ref{table:results_annotator-agreement}). Compared to the overall accuracy on the sample, the accuracy is higher with opposite filtering for all dimensions (i.e., complex and simple) and setups (i.e., triple and quadruple). The increase is higher for the quadruple than for the triple setup.}
\end{table}

\newpage
\paragraph*{Additional Annotation Results.} To further look at the difference between the quadruple and triple setup, we display additional results in Table \ref{fig:opposite-filter}. Here, we only consider the potential task instances that were correctly annotated in the opposite setup. For example, we take the task instances that were correctly annotated in the quadruple setup and see how many of them were also correctly annotated in the triple setup (in this case 0.68). One goal of this analysis was to see whether we can use annotations from the quadruple setup to also remove the ambiguities from the triple setup (or the other way around). However, in both cases accuracy is somewhat low (i.e., below 0.9) and we decided against such an approach. See Table \ref{table:error-analysis_annotation} for examples of every combination of (in)correctly annotated triple and quadruple setup of a potential task instance.

\begin{table*}[tp]
\centering \small
\begin{tabular}{p{12mm}| c p{28mm} p{28mm} p{28mm} p{28mm}}
\toprule
\textbf{Comp} & \textbf{Order}  & \textbf{Anchor 1 (A1)} & \textbf{Anchor 2 (A2)} & \textbf{Sentence 1 (S1)}& \textbf{Sentence 2 (S2)} \\
\midrule
\textbf{formal/\-informal} & S1-S2 & They were engaging in intercourse. & They were having sex. & You do not have the perspective. &  It's cause ya got no sense. \\
\midrule
\textbf{formal/\-informal} & S2-S1 & OH, REALLY? & Oh, is that so? & Girlfriends is one of my favorite shows on television. &  GIRLFRIENDS IS ONE OF MY FAVORITE SHOWS. \\
\midrule
\textbf{simple/\-complex} & S1-S2 & Many species had vanished by the end of the nineteenth century. & Many animals had disappeared by the end of the 1800s. & They are culturally akin. & Their culture is like the other.  \\
\midrule
\textbf{simple/\-complex}& S1-S2 & This stamp remained the standard letter stamp for the remainder of Victoria's reign, and vast quantities were printed. & This stamp stayed the standard letter stamp for the remainder of Victoria's reign, and a lot of them were printed. & Both names became defunct in 2007 when they were merged into The National Museum of Scotland. &  Both names stopped being used in 2007 when they became a part of The National Museum of Scotland. \\
\midrule
\textbf{number substitution} & S2-S1 & You are a n00b. & You are a noob. & This is cool. & This is c00l.  \\
\midrule
\textbf{number substitution} & S1-S2 & $\vert$-$\vert$0w n3rdy d0 y0u 1!k3 !7¿ ! d0 h4v3 4 107 0f !7 ;-) & How nerdy do you like it? I do have a lot of it ;-) & lol iM N0t CH3At1ng! & lol iM Not CHeating! \\
\midrule
\textbf{Shakes\-peare} & S1-S2 & Why, uncle tis a shame. & It's a shame, uncle. & O, wilt thou leave me so unsatisfied? & Oh, you're gonna leave me unsatisfied, right?  \\
\midrule
\textbf{formal/\-informal} & S2-S1 & i got limewire if i download songs on it will i get a ticket???  & Will I get a ticket if I download songs?  & The original song is very good. & The original song is like too good.....   \\
\midrule
\textbf{formal/\-informal} & S2-S1 & I like the Click Five and enjoy their songs. & The Click Five...they totally rock!their songs are out of this world!!  & i play guitar and some piano .....yet i cant read a note of music....lol & I can not read music, but I can play guitar and piano.   \\
\hline
\textbf{formal/\-informal} & S2-S1  & It reminds me of an old song from the Beatles. & Reminds me of an old beatles song... cant remember which one tho. & KEVIN n nfnhfnigbubjbni.....I dunt really watch American Idol.......... & Kevin, I am not exactly an `American Idol' viewer.  \\
\bottomrule
\end{tabular}
\caption{\label{table:screen_questions}
\textbf{Screening Questions.} List of manually created screening questions to test annotator quality. Anchor 2 is only used in the quadruple setup. The task is to match sentence 1 or sentence 2 with anchor 1 and anchor 2 to the respective other sentence. The correct matching is given in the Order column. The Shakespeare example was taken from \newcite{krishna-etal-2020-reformulating}. The rest were either inspired or taken from \newcite{xu-etal-2016-optimizing},  \newcite{rao-tetreault-2018-dear} and \newcite{Baumgartner_Zannettou_Keegan_Squire_Blackburn_2020}. 
}
\end{table*}

\begin{table*}
\centering \scriptsize %
\begin{tabular}{p{1mm}  p{1mm} p{4mm} p{6mm} | p{7mm} p{25mm} p{25mm} p{25mm} p{25mm} }
\toprule
\textbf{T} & \textbf{Q} & \textbf{Dim} & \textbf{Share} & \textbf{Order} & \textbf{Anchor 1} & \textbf{Anchor 2} & \textbf{Sentence 1}& \textbf{Sentence 2}  \\
\midrule
\xmark & \cmark & \textbf{for\-mal} &  $59 \approx 0.196$ & S1-S2 & List your best April Fools Pranks here & Please compile a list on here of your best April Fool pranks. & becuase in one of her songs she talks about saying no to sex pressure from her boyfriend &  In one of her songs ,she addresses the issue of not letting her boyfriend pressure her into having sexual intercourse.  \\
\midrule
\xmark & \cmark & \textbf{com\-plex} &  $94 \approx 0.312$  & S1-S2 & The Book of Nehemiah is a book of the Hebrew Bible, historically seen as a follow-up to the Book of Ezra, and is sometimes called the second book of Ezra. & The Book of Nehemiah is a book of the Hebrew Bible, historically regarded as a continuation of the Book of Ezra, and is sometimes called the second book of Ezra. & All the bats look up to him, and he says he caught two tiger moths which everyone in the colony knows to be a difficult feat for such a young bat & All the bats admire him, and he claims to have caught two tiger moths which are known by all the others in the colony to be an extraordinary achievement by such a young bat.\\
\midrule
\cmark & \xmark & \textbf{for\-mal} &  $14 \approx 0.047$ & S2-S1 & pointsreaper is lame he cannot sue Yahoo for him cheating, what a cry baby & He is not smart. You can not sue a website because you cheated. &  A woman did not perform the vocals. & A girl did not sing it. \\
\midrule
\cmark & \xmark & \textbf{com\-plex} &  $42 \approx 0.14$  & S1-S2 & Meanwhile the KLI has about 20 of those former Beginners' Grammarians. & Meanwhile, the KLI has about 20 of those past Beginner's Grammarians. & N-Dubz are a MOBO award winning hip hop group from London, based around Camden Town. & N-Dubz is a MOBO award winning hip hop group, based around Camden Town in London. \\
\hline
\xmark & \xmark & \textbf{for\-mal} &   $20 \approx 0.066$ & S1-S2 & Gentleman, and I thank God everyday for the one that I have! & I thank God for each day that I have. &  GIRLFRIENDS IS ONE OF MY FAVORITE SHOWS. & Girlfriends is one of my favorite shows on television.\\
\midrule
\xmark & \xmark & \textbf{com\-plex} &  $54 \approx 0.179$  & S1-S2 & Among the casualties were two fishers who were reported missing. & Two fisherman are missing among the people who may have been hurt or killed. & Baduhennna is solely attested by Tacitus' Annals where Tacitus records that a grove in Frisia was dedicated to her, and that near this grove 900 Roman prisoners were killed in 28 CE. & In Tacitus' Annals by Tacitus, it is recorded that a grove in Frisia was dedicated to Baduhennna, and near to this grove 900 Roman prisoners were killed in 28 CE. \\
\midrule
\cmark & \cmark & \textbf{for\-mal} &  $208 \approx 0.691$ & S1-S2 & im pretty sure that it was kiss & I am fairly certain it was a kiss. & Law and Order... it just has a clunk clunk & I like Law and Order, although it is a bit clunky lately. \\
\midrule
\cmark & \cmark & \textbf{com\-plex} &  $111 \approx 0.369$  & S1-S2 & Mifepristone is a synthetic steroid compound used as a pharmaceutical. & Mifepristone is a synthetic steroid compound which is used as a medicine. & The video was released on 7/14/06. & The video was premiered on MTV2 on July 14, 2006.\\
\bottomrule
\end{tabular}
\caption{\label{table:error-analysis_annotation}
\textbf{Annotation analysis.} For the simple/complex and the formal/informal dimensions, we give the number of occurrences of each combination of correct (\cmark) and wrong (\xmark) annotations in the triple (T) and quadruple (Q) setting. For every combination and style dimension an example is given. The share is calculated out of 301 examples. In total 602 examples were annotated for both Q and T settings with 301 per style dimension. The most common cases are \cmark \cmark and the \xmark \cmark combination for both style dimensions totaling 68.1\% and 88.7\% of the cases for the simple/complex and formal/informal dimensions respectively. There are ambiguous examples, where one could argue for both possible orders. After manual inspection, this seems to be more prevalent for the simple/complex dimension but it also happens for the formal/informal style dimension. E.g., for row (\xmark\xmark, formal), Anchor 1 could be understood as more formal (e.g., `gentleman') or more informal (e.g., `!' and an unusual grammatical structure). Row (\xmark\cmark, formal) is an example of the `triple problem'.
}
\end{table*}

\newpage

\section{Additional STEL Results}

In Table \ref{table:results-stel-random}, we display the share of task instances where models and methods could not decide between the two possible answers. This is adding more detail to the `random' column of Table \ref{table:results-stle-models}. The share of random decisions is lower for the more complex style dimensions (formal/informal: 0.05 and simple/complex: 0.13) and higher for the simpler style characteristics (nb3r substitution: 0.38 and contrac'tion usage: 0.15). This aligns with the intuition that the difference between the sentence pairs in the nb3r and contrac'tion dimension is smaller. The neural methods have a lower share of random decisions overall.

\begin{table*}[hbt!]
    \centering \small
    \begin{tabular}{ l | l l | l l | l l | l | l }
     \toprule
       & \multicolumn{2}{c|}{all} & \multicolumn{2}{c|}{formal} & \multicolumn{2}{c|}{complex} & nb3r & c'tion \\
        & filter & full & filter & full & filter & full & &\\
     \midrule
     BERT uncased & 0 & 0 & 0 & 0 & 0 & 0  & 0 & 0 \\ 
     BERT cased     & 0 & 0 & 0 & 0  & 0 & 0 & 0 & 0 \\
     RoBERTa        & 0 & 0 & 0 & 0  & 0 & 0 & 0 & 0 \\
     SBERT mpnet    & 0 & 0 & 0 & 0 & 0 &  0 & 0 & 0 \\
     SBERT para-mpnet & 0 & 0 & 0 & 0 & 0 & 0 & 0 & 0 \\
     USE            & 0.00 & 0.00 & 0.00 & 0.00   & 0.00 & 0.01 & 0 & 0 \\  %
     \midrule     
     BERT uncased NSP      & 0.10 & 0.12 & 0   & 0 & 0.21  & 0.21 & 0  & 0.19 \\  %
     BERT cased NSP   & 0.02 & 0.02 & 0 & 0 & 0.04  & 0.04 & 0  & 0.03 \\  %
     \midrule
     char 3-gram    & 0.05 & 0.05 & 0.03 & 0.04 & 0.01  & 0.01 & 0.57  & 0.02 \\ %
     word length    & 0.08 & 0.08 & 0.04 & 0.05 & 0.04  & 0.04 & 0.91  & 0 \\ %
     punctuation    & \textbf{0.38} & \textbf{0.39} & \textbf{0.31} & \textbf{0.31} & \textbf{0.42}  & \textbf{0.42} & \textbf{0.97} & 0.06 \\   %
     \midrule
     LIWC           & 0.09 & 0.09 & 0.01 & 0.01 & 0.12  & 0.13 & 0.53 & 0 \\ %
     LIWC (style)   & \textbf{0.62} & \textbf{0.64} & \textbf{0.37} & \textbf{0.38} & \textbf{0.80}  & \textbf{0.79} & \textbf{0.94} & \textbf{1.0} \\ %
     LIWC (function) & 0.28 & 0.28 & 0.14 & 0.14 & 0.38  & 0.38 & 0.81 & 0 \\ %
     deepstyle      & {0} & 0 & 0 & 0 & 0  & 0 & 0  & 0 \\     
     \midrule
     POS Tag        &  0.20 & 0.20 & 0.02 & 0.02 & 0.24  & 0.24 & 0.64  & \textbf{1.0} \\ %
     share cased    & 0.08 & 0.08 & 0.02 & 0.02 & 0.05  & 0.05 & 0.91  & 0 \\ %
     edit dist      & 0.08 & 0.07 & 0.01 & 0.01 & 0.05  & 0.05 & 0.52 & \textbf{0.33} \\ %
     \midrule
     \midrule
     Average & 0.11 & 0.11 & 0.05 & 0.05 & 0.13 & 0.13 & 0.38 & 0.15 \\
     \bottomrule
    \end{tabular}
    \caption{\textbf{Share of Random Decisions.} The share of task instances for which a method can not decide between the two options and decides randomly is given per dimension. The performance on the set of task instances before (full) and after crowd-sourced filtering (filter) is displayed. The two highest shares of random decisions is boldfaced. The share of random decisions is highest for the nb3r and lowest for the formal dimension. LIWC (style) and punctuation similarity have the overall highest share of random decisions.
     }
    \label{table:results-stel-random}
\end{table*}

\clearpage

\section{Task Generation}

\subsection{Contraction dictionary} 
The Wikipedia style guide discourages contraction usage and provides a dictionary with contractions that should be avoided.\footnote{\url{https://en.wikipedia.org/wiki/Wikipedia:List_of_English_contractions}} Some of those contractions are more colloquial (e.g., 'twas or ain't). We use an adapted version removing colloquial and less common contractions:  \{  
    ``aren't'': ``are not'',
    ``can't'': ``cannot / can not'',
    ``could've'': ``could have'',
    ``couldn't'': ``could not'',
    ``didn't'': ``did not'',
    ``doesn't'': ``does not'',
    ``don't'': ``do not'',
    ``everybody's'': ``everybody is'',
    ``everyone's'': ``everyone is'',
    ``hadn't'': ``had not'',
    ``hasn't'': ``has not'',
    ``haven't'': ``have not'',
    ``he'd'': ``he had / he would'',
    ``he'll'': ``he will'',  
    ``he's'': ``he has / he is'',
    ``here's'': ``here is'',
    ``how'd'': ``how did / how would'',
    ``how'll'': ``how will'',
    ``how's'': ``how has / how is'', 
    ``I'd'': ``I had / I would / I should'',
    ``I'll'': ``I shall / I will'',
    ``I'm'': ``I am'',
    ``I've'': ``I have'',
    ``isn't'': ``is not'',
    ``it'd'': ``it would / it had'', 
    ``it'll'': ``it shall / it will'',
    ``it's'': ``it has / it is'',
    ``mightn't'': ``might not'',
    ``mustn't'': ``must not'',
    ``must've'': ``must have'',
    ``needn't'': ``need not'',
    ``oughtn't'': ``ought not'',
    ``shan't'': ``shall not'',
    ``she'd'': ``she had / she would'',  
    ``she'll'': ``she shall / she will'',
    ``she's'': ``she has / she is'',
    ``should've'': ``should have'',
    ``shouldn't'': ``should not'',
    ``somebody's'': ``somebody has / somebody is'',
    ``somebody'd'': ``somebody would / somebody had'',  
    ``somebody'll'': ``somebody will'',
    ``someone's'': ``someone has / someone is'',
    ``someone'd'': ``someone would / someone had'',  
    ``someone'll'': ``someone will'',
    ``something's'': ``something has / something is'',
    ``something'd'': ``something would / something had'', 
    ``something'll'': ``something will'',
    ``that'll'': ``that will'', 
    ``that's'': ``that has / that is'',
    ``that'd'': ``that would / that had'',
    ``there'd'': ``there had / there would'',
    ``there'll'': ``there shall / there will'',
    ``there's'': ``there has / there is'',
    ``there've'': ``there have'',
    ``these're'': ``these are'',
    ``they'd'': ``they had / they would'', 
    ``they'll'': ``they shall / they will'',
    ``they're'': ``they are'', 
    ``they've'': ``they have'',
    ``wasn't'': ``was not'',
    ``we'd'': ``we had / we would / we should'', 
    ``we'll'': ``we shall / we will'',
    ``we're'': ``we are'',
    ``we've'': ``we have'',
    ``weren't'': ``were not'',
    ``what's'': ``what has / what is / what does'',
    ``when's'': ``when has / when is'',
    ``who'd'': ``who would / who had'', 
    ``who'll'': ``who will'',  
    ``who's'': ``who has / who is'', 
    ``won't'': ``will not'',
    ``would've'': ``would have'',
    ``wouldn't'': ``would not'',
    ``you'd'': ``you had / you would'', 
    ``you'll'': ``you shall / you will'',
    ``you're'': ``you are'',
    ``you've'': ``you have''\}

\subsection{Number substitutions} We selected a pool of potential sentences where words contained character substitution symbols ({4,3,1,!,0,7,5}) or  are part of a manually selected ``seed list'' of number substitution words{\footnote{Inspired by \url{https://www.gamehouse.com/blog/leet-speak-cheat-sheet/}, \url{https://simple.wikipedia.org/wiki/Leet}, \newcite{gr8_to_great}, \url{ https://h2g2.com/edited_entry/A787917} and manually looking at a few Reddit posts}}: \\
\{ ``2morrow'': ``tomorrow'', 
    ``c00l'': ``cool'', 
    ``n!ce'':``nice'',
    ``l0ve'':``love'',
    ``sw33t'':``sweet'',
    ``l00k'':``look'',
    ``4ever'':``forever'',
    ``l33t'':``leet'',
    ``1337'':``leet'',
    ``sk8r'':``skater''
    ``n00b'':``noob'',
    ``d00d'':``dude'',
    ``ph34r'':``fear'',
    ``w00t'':``woot'', 
    ``b4'':``before'',
    ``gr8'':``great'',
    ``2day'':``today'',
    ``t3h'':``teh'', 
    ``m4d'':``mad'',
    ``j00'':``joo'',
    ``0wn'':``own'',
    ``h8'':``hate'',
    ``w8'':``wait''
\}
    
Then, we manually filtered out sentences without number substitutions (e.g., common measuring units or product numbers). Our resulting list of 100 sentences pairs contains more substitution words than the above ``seed list'' (e.g., ``d4rk'', ``appreci8'', ``h1m'').

\clearpage
\section{Similarity-based Decision}

\begin{figure}[h!]
    \centering
    \includegraphics[width=0.5\textwidth]{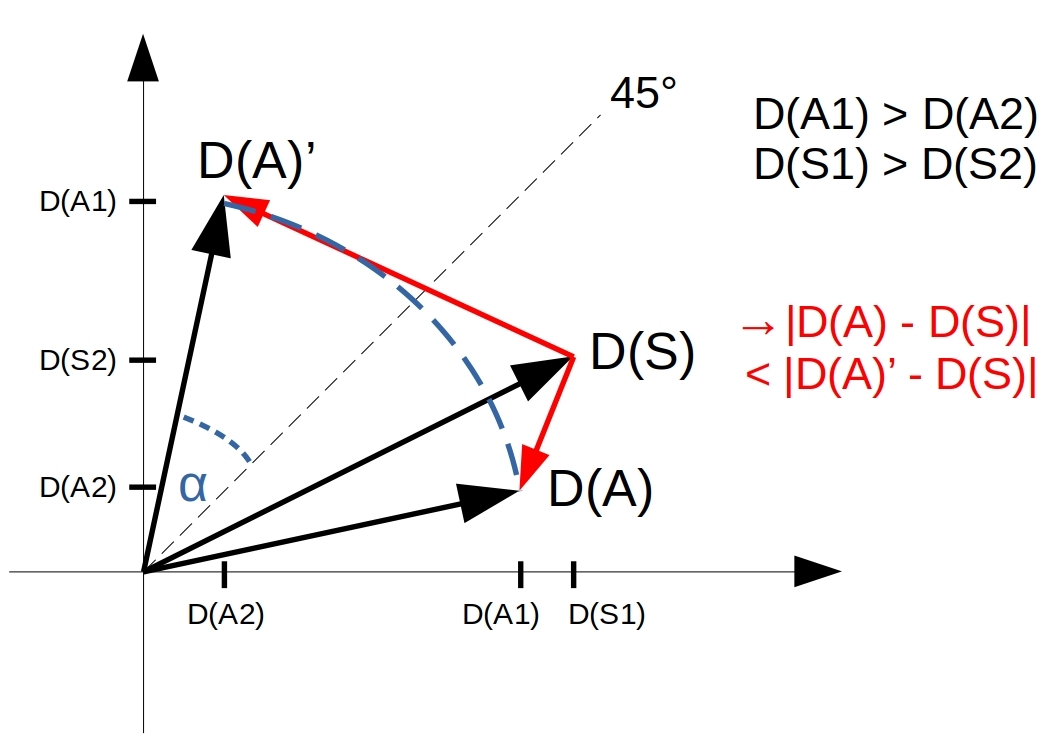}
    \caption{\textbf{Proof Sketch.}
    Let $D$ be the considered style component (e.g., formal/informal) and $D(\text{A}1)$, $D(\text{A}2)$, $D(\text{S}1)$, $D(\text{S}2)$ be the localization of $\text{A}1, \text{A}2, \text{S}1, \text{S}2$ along that component. W.l.o.g., let the correct ordering be $\text{S}1$-$\text{S}2$ and $D(\text{A}1)>D(\text{A}2)$. %
    Let us assume that for all other style and content aspects $\widetilde{D}$ (e.g., simple/complex), $\widetilde{D}(\text{A}1)=\widetilde{D}(\text{A}2)$ and $\widetilde{D}(\text{S}1)=\widetilde{D}(\text{S}2)$ hold. %
    We define $D(\text{A}):=\begin{pmatrix}D(\text{A}1) & D(\text{A}2)\end{pmatrix}^\intercal$ and $D(\text{S}):=\begin{pmatrix}D(\text{S}1) & D(\text{S}2)\end{pmatrix}^\intercal$ as the style vectors of the combined anchor ($\text{A}1$ and $\text{A}2$) and alternative sentences ($\text{S}1$ and $\text{S}2$).  %
       Then, with the correct ordering being $\text{S}1$-$\text{S}2$ and $D(\text{A}1)>D(\text{A}2)$,  $D(\text{S}1)>D(\text{S}2)$ holds. Thus, both $D(\text{A})$ and $D(\text{S})$ point to a coordinate below the $45^\circ$-axis when the first component of the respective vectors corresponds to the $x$-axis and the second to the $y$-axis (see sketch).
    Let $D(\text{A})'$ be the reflected vector of $D(\text{A})$ along the $45^\circ$-axis, i.e., $\begin{pmatrix}D(\text{A}2) & D(\text{A}1)\end{pmatrix}^\intercal$.
    Then, as shown in the sketch, the length of the vector $D(\text{A})-D(\text{S})$ is smaller than $D(\text{A})'-D(\text{S})$ as the angle between $D(\text{A})$ and $D(\text{S})$ will always be smaller than the one between $D(\text{A})'$ and $D(\text{S})$.
    This corresponds to equation (1), when replacing similarities (i.e., $1-\text{sim}(x,y)$) with distances (i.e., $|x-y|$).  Thus, equation (1) holds when working with style-sensitive similarity functions that can be translated to distances. Note: As only cosine `angular distance' is a distance metric, this would need to be the angular cosine similarity. However, angular cosine similarity can be replaced by cosine similarity in inequality (1) as relative ordering is the same for the two similarity metrics. %
    }
    \label{fig:model-dist}
\end{figure}

\section{Computing Infrastructure}
The evaluation of the 18 (language) models and methods took 14 hours in total on a machine with 32 GB RAM and 8 intel i7 CPUs using Ubuntu 20.04 LTS. No GPU was used.

\end{document}